\documentclass[runningheads]{llncs}
\usepackage[justification=justified, labelfont=bf, format=plain]{caption}
\usepackage{subfigure}
\usepackage{amsmath,amssymb}
\usepackage{color}
\usepackage[colorlinks]{hyperref}
\usepackage{amsmath}
\usepackage{booktabs} 
\usepackage{multirow}
\usepackage{float}
\usepackage[dvipsnames]{xcolor}
\usepackage{footnote}
\usepackage{tabularx}
\usepackage{cite}
\usepackage{soul}
\usepackage{xurl}
\usepackage{graphics, graphicx}
\usepackage{adjustbox}
\usepackage{bm}
\usepackage{tablefootnote}
\usepackage{threeparttable}
\usepackage{setspace}
\usepackage{algorithm2e}

\PassOptionsToPackage{hyphens}{url}
\DeclareMathAlphabet{\pazocal}{OMS}{zplm}{m}{n}

\definecolor{purple}{RGB}{111, 4, 194}
\definecolor{green}{RGB}{6, 161, 6}
\definecolor{orange}{RGB}{255, 140, 0}

\DontPrintSemicolon 
\RestyleAlgo{ruled} 
\SetKwComment{Comment}{/* }{ */}

\title{Enhancing Few-Shot Image Classification\\with Cosine Transformer}
\titlerunning{Enhancing Few-Shot Image Classification with Cosine Transformer}
\authorrunning{Quang-Huy Nguyen et al.}
\date{October, 2022}
\author{Quang-Huy Nguyen\inst{1, 2} \and
Cuong Q. Nguyen \inst{1} \and\\
Dung D. Le\inst{2} \and 
Hieu H. Pham\inst{1, 2} }
\institute{
VinUni-Illinois Smart Health Center, VinUniversity, Hanoi, Vietnam;
\and
College of Engineering \& Computer Science, VinUniversity, Hanoi, Vietnam;\\ 
\email{\{huy.nq,cuong.nq,hieu.ph,dung.ld\}@vinuni.edu.vn}
}

\begin{document}
\maketitle
\setstretch{1.1}
\begin{abstract}
This paper addresses the few-shot image classification problem, where the classification task is performed on unlabeled query samples given a small amount of labeled support samples only. One major challenge of the few-shot learning problem is the large variety of object visual appearances that prevents the support samples to represent that object comprehensively. This might result in a significant difference between support and query samples, therefore undermining the performance of few-shot algorithms. In this paper, we tackle the problem by proposing Few-shot Cosine Transformer (FS-CT), where the relational map between supports and queries is effectively obtained for the few-shot tasks. The FS-CT consists of two parts, a learnable prototypical embedding network to obtain categorical representations from support samples with hard cases, and a transformer encoder to effectively achieve the relational map from two different support and query samples. We introduce Cosine Attention, a more robust and stable attention module that enhances the transformer module significantly and therefore improves FS-CT performance from 5\% to over 20\% in accuracy compared to the default scaled dot-product mechanism. Our method performs competitive results in \textit{mini}-ImageNet, CUB-200, and CIFAR-FS on 1-shot learning and 5-shot learning tasks across backbones and few-shot configurations. We also developed a custom few-shot dataset for Yoga pose recognition to demonstrate the potential of our algorithm for practical application. Our FS-CT with cosine attention is a lightweight, simple few-shot algorithm that can be applied for a wide range of applications, such as healthcare, medical, and security surveillance. The official implementation code of our Few-shot Cosine Transformer is available at \url{https://github.com/vinuni-vishc/Few-Shot-Cosine-Transformer}.\footnote{This paper is published at IEEE Access with DOI: \href{https://doi.org/10.1109/ACCESS.2023.3298299}{10.1109/ACCESS.2023.3298299}. Please visit the official version of this paper at IEEE Xplore for appropriate citation.}
\end{abstract}

\section{Introduction}

Deep learning methods have shown promising performances in numerous computer vision tasks and real-world applications by leveraging large-scale annotated data \cite{deng2009imagenet, simonyan2014very}. However, collecting and labeling an adequate amount of data for training such methods is extremely costly and time-consuming, particularly in some special contexts such as medical treatment, healthcare monitoring, or security surveillance, where data availability is limited due to scarcity, expensive, or privacy concerns arising respectively. In many practical applications, the balance between quality and quantity for building datasets often be considered for the optimal cost of training and maintenance. For example, developing a smart system that can simultaneously monitor body movement, heart rate, and respiration of yoga learners effectively requires building an extremely costly dataset annotated by various tracking devices under the supervision of specialists in different domains, making by far large-scale dataset development impractical. To address such data limitation challenges in deep learning, few-shot image classification \cite{vinyals2016matching,chen2019closerfewshot,li2023deep} is proposed as one solution for learning to predict unseen data with a very restricted quantity of labeled instances.

In a few-shot learning scenario, typically, the feature representation network is trained with a constrained amount of labeled \textit{support} data to classify unlabeled \textit{query} instances sharing the same categories \cite{finn2018probabilistic,wang2020generalizing, zhang2021shallow}. This approach enables the model to recognize images with only a small quantity of data. To evaluate the few-shot learning model effectively, the training set and the testing set are disjoint in terms of categories and divided into episodic tasks. From a learning perspective, few-shot learning can be categorized into two approaches: (i) \textit{inductive learning} \cite{shao2021mhfc, zheng2022subgraph, shin2022active} that categorizes query instance independently by exploiting per query instance representation for decision making and (ii) \textit{transductive learning} \cite{qiao2019transductive,masud2021transductive,zhu2023transductive} that categorizes query instances concurrently, exploring the entire relationship between support and query instances within a task. From an algorithmic perspective, few-shot learning can be divided into two strategies: (i) learning an embedding space to cluster data features by category under a distance-based metric \cite{ sung2018learning, wang2018low, afrasiyabi2022matching} and (ii) fine-tuning model's parameters effectively for a novel task with only a few gradient steps \cite{ravi2016optimization, sun2019meta, hu2022pushing}.

While few-shot learning methods have shown promise in dealing with limited data problems, they still face two critical challenges that significantly affect their learning capacity. First, the limited quantity of support instances may not fully represent the target categories under various visual conditions such as appearance, point of view, and object shape. As the support instances are often chosen randomly, the selected samples might only focus on the hard cases and do not adequately capture the characteristics of the target object. Thus, effectively handling hard samples is crucial in few-shot learning. Second, the support and query distribution might differ greatly due to random sampling and extremely small sample sizes. Therefore, it is crucial to establish a meaningful relationship between these two sets, particularly in transductive learning approaches, where aligning the support and query distributions becomes essential.

To tackle these challenges, we have developed a novel framework aimed at improving support representations and enhancing the transductive few-shot classification process. Our framework consists of two main modules: enhanced prototypical embedding and transformer encoder with a cross-attention mechanism for effective support-query relational mapping. Firstly, the prototypical embedding is enhanced with learnable and adjustable mean weights for computing centroids. This ensures that the prototypical representations take into account both easy and hard samples, resulting in more comprehensive embeddings. Secondly, we introduce a novel Cosine Attention mechanism based on cosine similarity, replacing the traditionally scaled dot-product one. This cosine attention eliminates the disparities between two feature sets, leading to a more stable and effective relational map. It significantly highlights the correlation between support and query samples that share the same characteristic. The correlation map is then used for query prediction in a transductive learning manner. These two improvements on the baseline framework enhance the few-shot image classification methods, resulting in our proposed method \textbf{F}ew-\textbf{s}hot \textbf{C}osine \textbf{T}ransformer (FS-CT).

Through a comprehensive and empirical evaluation, we demonstrate the effectiveness of our proposed FS-CT under various configurations and datasets. We further analyze the effectiveness of the proposed cosine attention under various transformer-based few-shot algorithms in ablation studies. These studies validate the impact of our improvements for few-shot image classification. To showcase the real-world applicability of our method, we developed a small-scale custom dataset specifically for yoga poses recognition. This dataset consists of 50 categories and nearly 2,500 images. The performance results of FS-CT on this Yoga dataset showcase the potential of few-shot learning in healthcare practical applications in general, and smart Yoga monitoring systems in particular. The official implementation code for FS-CT along with the proposed Yoga Poses Recognition dataset are available at \url{https://github.com/vinuni-vishc/Few-Shot-Cosine-Transformer}.

Our contributions can be summarized as follows:
\begin{itemize}
    \item We propose a novel few-shot image classification method called Few-shot Cosine Transformer (FS-CT), which incorporates a prototypical embedding module and a Transformer encoder architecture.
    \item We improve the conventional prototypical embedding by introducing a learnable weighted mean operation. This helps enhance the categorical representations for the support set and mitigates the impact of hard samples in few-shot tasks.
    \item We develop a new attention mechanism called cosine attention, which enhances the transformer layer's ability to map support and query features. These results in a more stable and significant correlation map, leading to improved transformer outputs and more accurate few-shot predictions.
    \item We demonstrate the effectiveness of our proposed FS-CT method, particularly in conjunction with the improved cosine attention mechanism through detailed empirical evaluations and ablation studies across few-shot datasets.
    \item We develop a custom few-shot dataset for the Yoga pose recognition task, where our method's performance showcases the potential of few-shot learning in practical healthcare applications, particularly in Yoga scoring and monitoring system.
\end{itemize}

The rest of this paper is organized as follows. Section \ref{sec:Related} reviews the previous studies. Section \ref{sec:Fomulation} provides a mathematical formulation of the few-shot classification problem and Section \ref{sec:Method} specifies the proposed FS-CT with Cosine Attention mechanism. Section \ref{sec:Experiment} presents our experimental settings and evaluation. Section \ref{sec:Discussion} discusses limitations and future works. Section \ref{sec:Conclusion} concludes the paper.

\section{Related Work}
\label{sec:Related}
The main objective of this study is to enhance the accuracy of few-shot image classification in a transductive learning setting by computing a correlational map between the support and query sets using transformer attention. This section provides a brief overview of the related research on the image classification method, few-shot learning, and the Transformer attention mechanism employed to compute the correlational map to support our proposed method.

\subsection{Image Classification}
Image classification is one of the fundamental tasks of computer vision, where deep models are developed to recognize images based on their content. Although general Deep learning methods have demonstrated their effectiveness in performing classification tasks on large-scale datasets with deep neural network architectures for most tasks \cite{deng2009imagenet, simonyan2014very, he2016deep}, image classification often faces challenges in specific domain areas or suffering under various conditions, e.g. imbalance data, or data in small scales. Therefore, improvements for deep neural networks often be made to tackle the problem. For example, in dealing with small dataset problems, improvements often be made for network architecture, cost function, data augmentation, latent augmentation (adversarial training), and warm-starting method with pre-trained model \cite{brigato2022image}. Zhou et. al. \cite{zhou2018local} improved neural network by developing an entanglement coefficient algorithm between pixels based on quantum physics perspective as a general case of various traditional distance functions on geometric sensing images. Zhou et. al. \cite{zhou2022study} considered both local and global features for blind quality prediction of natural scene images without prior knowledge. Ban et. al. \cite{ban2022depth} focused on enhancing the quality of microscopic images captured by a monocular camera using depth estimation techniques. Their aim was to improve  the images before applying image recognition tasks. While these improvements often bring benefits for specific domain tasks, hyper-parameter optimization is currently underestimated and should be considered in future studies to ensure more accurate evaluations and fair comparisons between methods \cite{brigato2022image}.

Instead of exploring deeply the nature of images from various perspectives for domain-specific tasks, we focus in our research on enhancing network architectures to address small dataset challenges in few-shot learning for various domain tasks. Our technique is highly adaptable, as it can be constructed on any feature backbone architecture using pre-trained models. We conduct a thorough evaluation procedure for our proposed method and provide a fair comparison between ours and existing studies under very detailed experiments and ablation studies with an optimized hyper-parameter configuration.

\subsection{Few-shot Learning} 

Few-shot learning is a subset of meta-learning \cite{ ravi2016optimization,sun2019meta,kang2021relational}, which develops models that are able to adapt to unseen tasks with small training data. Meta-learning algorithms take an advance on prior knowledge from a large-scale dataset (\textit{e.g.}, a pre-trained deep network) to effectively learn on a small novel dataset via a \textit{meta-learner} (or \textit{few-shot learner}) \cite{finn2017model, fei2006one}. Based on the learning method, few-shot learning algorithms can be divided into two categories. (i) \textit{Metric-based learning} \cite{shin2022active, snell2017prototypical, doersch2020crosstransformers, qiao2018few, hou2019cross} focuses on learning an embedding space where samples from the same category are mapped closely together under a distance metric: Active Instance Selection \cite{shin2022active} fits categorical distribution for support set, and selects new instances for support set based on a clustering algorithm using a metric distance, thereby improving few-shot learning performance. This method relies on the assumption that the distribution of the support set can be well approximated by a specific model, which may not hold true in all scenarios. Prototypical Network \cite{snell2017prototypical} computes prototypical embeddings by averaging support features within the same category as prototypical embeddings, and then measures the Euclidean distance between queries and prototypical embeddings. However, it relies solely on averaging support features without considering the relative importance of different samples, leading to suboptimal representations, especially when dealing with hard or challenging samples. (ii) \textit{Optimization-based learning} \cite{chen2019closerfewshot,finn2018probabilistic, ravi2016optimization, hu2022pushing, lee2018gradient} fine-tune model's parameters to quickly adapt to the new task with only a few effective gradient descent steps: Baseline++ \cite{chen2019closerfewshot} improves the fine-tuning step by replacing the dot product operation with cosine similarity in the linear classifier layer. Meta-Learner LSTM \cite{ravi2016optimization} employs LSTM for updating and tracking parameters across few-shot tasks, enabling fast adaptation. P$>$M$>$F \cite{hu2022pushing} adopts a three-stage approach: pre-training the feature backbone on unlabeled external data, re-training the model using the prototypical network, and finally fine-tuning it on novel tasks with a few gradient steps while employing data augmentation. 

Among the two categories of few-shot learning methods, metric-based learning stands out as a straightforward yet effective approach. It typically comprises two stages: \textit{feature extraction} to obtain extracted features from both labeled and unlabeled samples under the same embedding space, and \textit{metric function} to utilize a similarity or distance metric for categorizing unlabeled samples by comparing \cite{sung2018learning} or clustering \cite{snell2017prototypical} the embedded features. Based on the inference settings, there are two learning approaches for the few-shot algorithm, including \textit{inductive learning} \cite{vinyals2016matching, shao2021mhfc, finn2017model} and \textit{transductive learning} \cite{sung2018learning, liu2018learning, bateni2022enhancing}. Inductive learning classifies each query sample individually, while transductive learning classifies every query sample collectively \cite{shin2022active, boudiaf2020information}. The latter learning method allows additional information in data distribution or visual resemblance can be obtained and leveraged among query samples, thus potentially improving the overall performance. 

In this study, we explore the alignment between support and query features for few-shot recognition in a transductive manner with a metric-based learning approach. We focus on the cross-attention mechanism in the transformer as an effective method for support-query correlation maps and discuss its advantages and limitations in the following section.

\subsection{Transformer Attention mechanism}
After being introduced in \cite{vaswani2017attention} for natural language processing tasks, transformer soon rose to dominance in computer vision \cite{dosovitskiy2020image,carion2020end,cheng2021per}. The core of the transformer is the attention mechanism, which calculates an attention map that indicates the similarity between features for solving tasks. The mechanism comes with two variants, \textit{self-attention} determines the internal relationships within a feature set, and \textit{cross-transformer} calculates the external relationships between two feature sets. Several few-shot learning studies are inspired by the transformer and its attention mechanism \cite{afrasiyabi2022matching, doersch2020crosstransformers, ye2020few, liu2020universal, han2022few}, where the methods, in general, involve attention mechanism to align labeled and unlabeled feature for classification. SetFeat \cite{afrasiyabi2022matching} tackles few-shot classification by matching support and query features at multiple scales using shallow attention mechanisms, incorporating various distance-based methods. CTX \cite{doersch2020crosstransformers} focuses on obtaining a coarse alignment between query and support samples by emphasizing local features through an improved spatial attention mechanism. URT \cite{liu2020universal} takes a different approach by computing a universal representation between labeled and unlabeled samples through the averaging of multiple scaled dot-product attention on domain-specific representation. While attention is a powerful mechanism, it has critical problems of missing good insight about attention and an expensive quadratic computational cost. Thus, research on the attention mechanism focuses on three main directions: (ii) reducing the computation cost, (ii) obtaining a good insight for attention, and (iii) designing a good attention mechanism for a specific task. Most studies on attention often focus on the self-attention mechanism with one set of features as input rather than two in cross-attention. Therefore, the difference between two feature sets that might happen in cross-attention becomes unnoticed. This made the attention output becomes unstable, thus reducing the transformer performance. 

In this study, we investigate this limitation inside the scaled dot-product attention mechanism and propose a replacement cosine attention to tackle the problem. Although prior research has utilized the cross-attention mechanism for few-shot learning, our work stands out as the first to investigate deeply the cross-attention limitation for the few-shot classification problem. To the best of our knowledge, this is the first time the cosine similarity-based attention mechanism has been explored and proven its effectiveness in the tasks of few-shot learning.

\section{Problem formulation and notations}
\label{sec:Fomulation}

We first formalize a standard few-shot classification problem while introducing some notations. In the few-shot learning problem, the objective is to develop a few-shot model that is able to perform tasks, in this case, image classification, on any set of random categories given only a very small amount of labeled samples per category as support information. Given a train set $D_{\textnormal{train}}$ and few-shot learner $A(.\mid\theta)$. The objective of few-shot classification is to learn the optimal parameter $\theta^*$ so that it can achieve a good performance of algorithm $A(.\mid\theta)$ on a test set $ D_{\textnormal{test}}$. $ D_{\textnormal{train}}$ and $ D_{\textnormal{test}}$ must be disjoined in categories. 

\begin{figure}[!ht]
    \centering
    \includegraphics[width=0.8\textwidth]{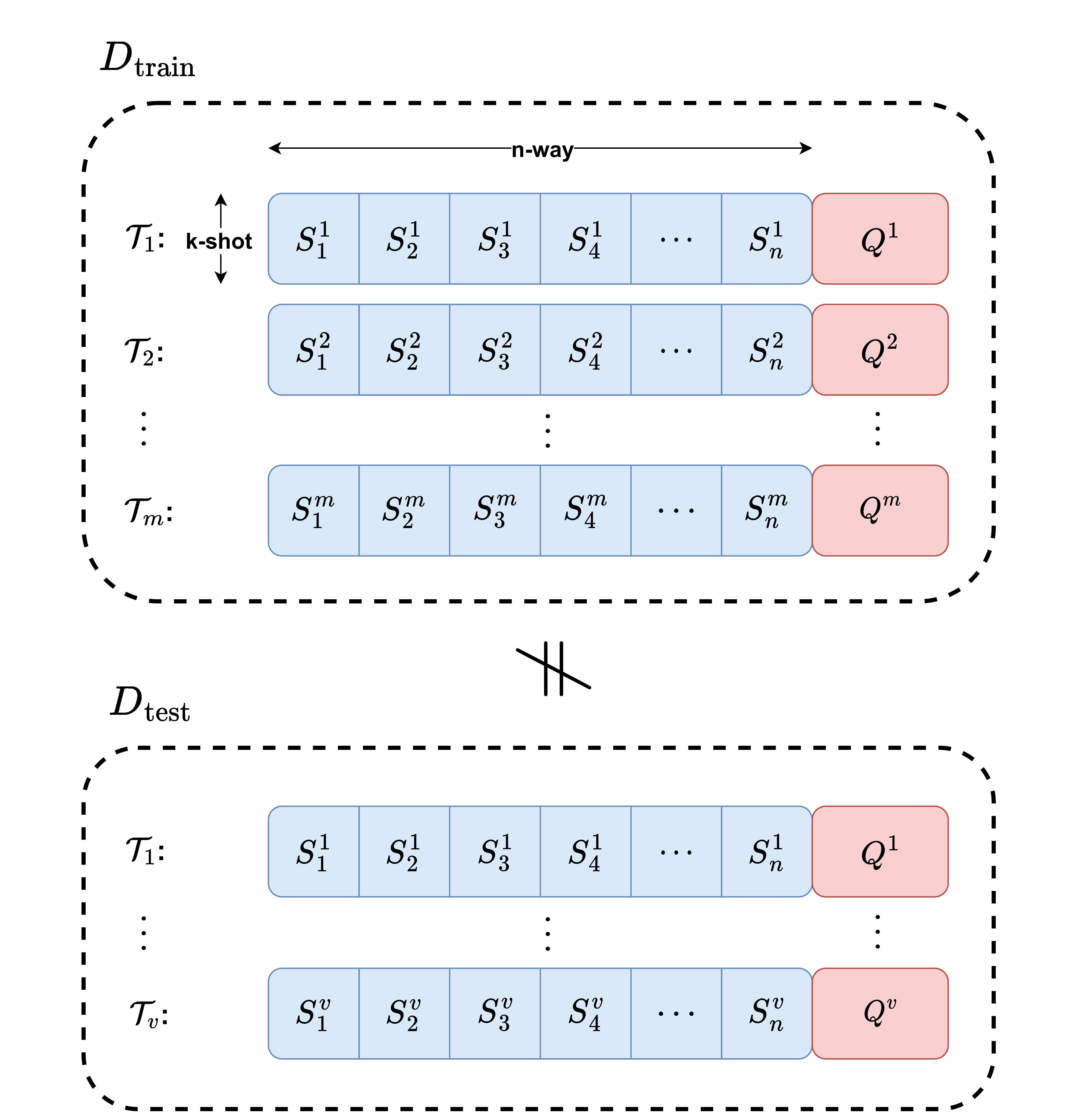}
    \caption{\textbf{Formulation of few-shot learning problem} including the training set $D_\text{train}$ with $m$ tasks and the testing set $D_\text{test}$ with $v$ tasks. Here, $m$ and $v$ could be equal or different. Each task $\mathcal{T}$ comes with different sets of categories and consists of labeled support set $S$ and unlabeled query set $Q$ that share the same categories. Support set $S$ follows the \textit{n}-way \textit{k}-shot setting. $D_\text{train}$ and $D_\text{test}$ are disjointed in categories. A \textit{few-shot learner} $A$ is trained on $D_\text{train}$ to perform test on $D_\text{test}$.}
    \label{fig:FS_Setting}
\end{figure}

A few-shot learning problem is usually trained with an episodic learning strategy, where the proposed approach is trained and tested on different tasks with different sets of categories. The episodic learning with $m$ tasks is generally described in Fig. \ref{fig:FS_Setting}. Individual task $\mathcal{T} = \{S,\ Q\} \sim p(D)$ is derived randomly from data set $D$, where $D$ can be either the training set $D_\text{train}$ or testing set $D_\text{test}$ depending on the training scheme and $p(D)$ is the distribution over $D$. Each task $\mathcal{T}$ consists of two sub-sets: labeled support set $S$ and unlabeled query set $Q$ that share the same $n$ ground truth categories. The objective is to train the few-shot model to be able to classify $Q$ given only $S$ as support information on any arbitrary $n$ categories. Task $\mathcal{T} = \{S,\ Q\}$ is also called \textit{episodic batch} or episode for short. One training epoch may contain various episodes.

\begin{figure}[!ht]
    \centering
    \begin{adjustbox}{max width=1.2\textwidth,center}
    \includegraphics[width=1.2\textwidth]{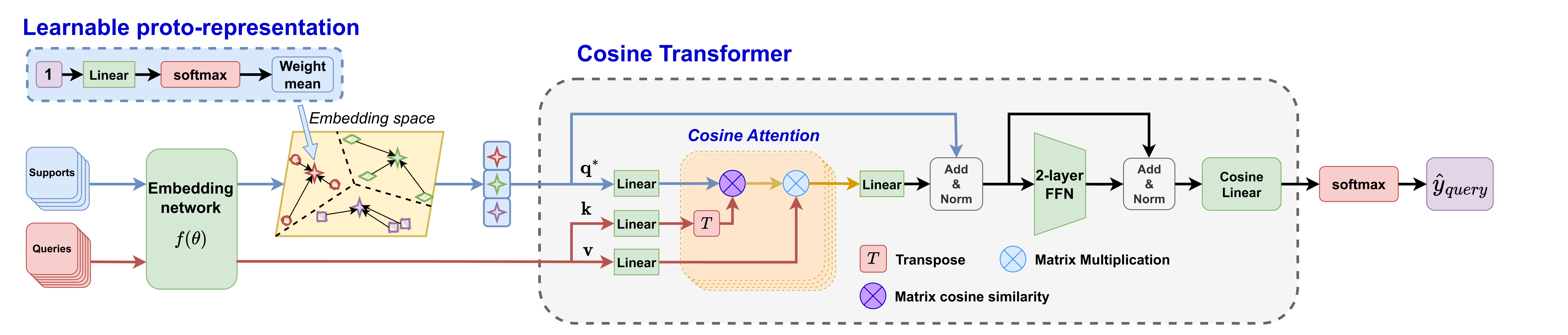}
    \end{adjustbox}
    \caption{\textbf{The overall architecture of the proposed Few-shot Cosine Transformer}, which includes two main components: (a) \textit{\textbf{learnable proto-representation}} that calculates the categorical proto representation given random support features that might be either in the far margin of the distribution or very close to each other and (b) \textit{\textbf{Cosine Transformer}} that determines the similarity matrix between proto representations and query samples for the few-shot classification tasks. The heart of the transformer architecture is \textit{\textbf{Cosine Attention}}, a cross-attention mechanism with cosine similarity and no softmax function to deal with two different sets of features. The Cosine transformer shares a similar architecture with a standard transformer encoder block, with two skip connections to preserve information, a two-layer feed-forward network, and layer normalization between them to reduce noise. The outcome value is through a cosine linear layer, with cosine similarity replacing the dot-product, before feeding to softmax for query prediction. This figure is better viewed in color.}
    \label{fig:Proposed_Cosine}
\end{figure}

Let $(x,\ y)$ be defined as the input image sample and its ground truth, respectively. The objective for a few-shot classification task is to predict labels of the query set $Q=\left\{\boldsymbol{x}_q^{i}\right\}_{i=1}^q$, given the support set  $S=\left\{(\boldsymbol{x}_s,\ y_s)^{i,j}\right\}_{i=1,\cdots,n}^{j=1,\cdots,k}$, where $y_s \in C$ for a set of $n$ categories (\textit{$n$-way}), $k$ is the number of training samples per category (\textit{$k$-shot}), and $q$ is the total number of $Q$ samples. The number of $k$ must be small in a few-shot setting. In this paper, we explore two configurations: 5-way 5-shot and 5-way 1-shot. The few-shot classification problem can be formulated with the following optimization formula:

\begin{equation}
    \theta^* = \underset{\theta}{\arg\min}\ \mathbb{E}_{(S,Q) \sim p(D)}\pazocal{L}\Big(S, Q;\ A(. \mid \theta)\Big),
    \label{equa:loss_func}
\end{equation}
where
\begin{equation*}
\begin{aligned}
    \pazocal{L}\Big(S, Q;\ A(. \mid \theta)\Big) = \ &\frac{-1}{q}\sum _{(\boldsymbol{x}_q,y) \in Q} \log \left[\ p\left(y \ \big|\ A(\boldsymbol{x}_q\mid\theta);\ S\right) \right] \\ &+ \lambda R(\theta),
\end{aligned}
\end{equation*}

with $p\left(y \ \big|\ A(\boldsymbol{x}_q\mid\theta);\ S\right) $ is the probabilistic prediction of sample $\boldsymbol{x}_q \in Q$ on true label $y$ using few-shot algorithm $A(.|\theta)$ given  for $\boldsymbol{x}_q$ and support set $S$. $\lambda R(\theta)$ is an optional regularization with factor $\lambda$. The loss function $\pazocal{L}\Big(S, Q;\ A(.|\theta)\Big)$ is dependent on the few-shot problem and method. In this work, the Categorical Cross-entropy loss Equation \ref{equa:loss_func} is explored for categorical classification.

\section{Few-Shot Cosine Transformer (FS-CT)}
\label{sec:Method}
In this section, we describe the proposed Few-Shot Cosine Transformer (FS-CT) architecture, which utilizes transformer framework to learn the similarities between labeled support  and unlabeled query features to recognize query samples following the transductive learning approach. Fig. \ref{fig:Proposed_Cosine} presents the overall architecture of FS-CT, with two main components: learnable prototypical embedding, and Cosine Transformer. 

Overall, FS-CT shares a similar architecture with the transformer encoder architecture. Given two input support set $S$ and query set $Q$, their images are fed into a backbone feature extractor to obtain two feature tensors $Z_S$ and $Z_Q$, and then features from $Z_S$ are averaged along individual categories to obtain the prototypical representation $Z_P$ like the prototypical network \cite{snell2017prototypical}. Unlike other conventional transformer-based architectures, positional encoding is removed, since the arrangement of features is unimportance. After that, $Z_P$ and $Z_Q$ are brought into three linear layers to split into a multi-head of three features $\langle\mathbf{q^\ast}$, $\mathbf{k}$, $\mathbf{v}\rangle$\footnote{To avoid the conflict from double usage of the term ``query'', we denote the queries $\mathbf{q^\ast}$ of the attention mechanism to distinguish with the query set $Q$ with sample size $q$ in the few-shot tasks.}, then go through a multi-head cross-attention mechanism to obtain the weight attention features of $Z_Q$ on $Z_P$. Instead of using the vanilla softmax attention, we propose a variation of the attention mechanism named ``\textit{cosine attention}'', which utilizes cosine similarity to calculate the attention weight. The outcome attention values between heads are then fed to an output linear layer to combine heads together, followed by a two-layer MLP with GELU activation function. Two skip-connected layers are applied to prevent losing information and layer normalization is applied before the linear layer for smoothing values throughout the FS-CT. Finally, the outcome feature is brought through a Cosine linear layer followed by the softmax to yield probabilistic scores on individual categories to predict queries' labels. In the following subsections, we will describe in detail the essential modules within our FS-CT method.

\subsection{Learnable Prototypical representation}
Given support set $S={(X_S,\ Y_S)}$ and query set $Q={(X_Q, Y_Q)}$ follows few-shot setting ($n$-way, $k$-shot, $q$ query samples) with $Y_S, \ Y_Q \in n$ categories; $X_S,\ X_Q \in \mathbb{R}^{c \times h \times w}$; $X_S$ and $X_Q$ are first fed to a backbone feature extraction $f(.\mid\theta)$ to obtain the feature representations $Z_S \in \mathbb{R}^{n \times k \times d}$, $Z_Q \in \mathbb{R}^{q \times d}$. Then, all $k$ learned features $\{z_c^i\} \in Z_S, \ i \in [1,k]$ that share the same category $c$ are then average equally (arithmetic mean) to obtain the prototypical representation $Z_P \in \mathbb{R}^{n \times d}$, with $z_c \in Z_P$ represented for the centroid of category $c$. However, with the few supporting samples chosen randomly, the prototypical representations are not guaranteed to be well represented using this arithmetic mean approach. This may happen when the embeddings of chosen samples are in the far margin of the categorical distribution space or close to each other as in Fig. \ref{fig:Proposed_Cosine}. This problem becomes critical when the number of shots is low. To tackle the problem, we propose an improvement to turn the arithmetic mean into \textit{mean as weighted sum} (or weighted mean), where the weights can be adjusted through the learning process to obtain a better prototypical representation for each category given the same small support samples. The formula for the learnable prototypical embedding is described by:

\begin{equation}
    z_P = \sum^k_{i=1} z^i_{s} \ . \ \textnormal{Softmax} \left[g(a\mid\theta_P)\right],
    \label{equal:Weight_Averaging}
\end{equation}

Initially, the mean weights $W_{avg}=g(a\mid\theta_P)$ for calculating categories' centroid are obtained by feeding a fixed scalar input $a$ into the linear layer $g(.\mid\theta_P)$ with learnable weight $\theta_P \in \mathbb{R}^{n \times k \times 1}$ and no bias. We used $a = 1$ so that the output $W_{avg}$ can have the same value with the weight $\theta_P$. After achieving the mean weight, Softmax is applied along the $k$ axis to ensure that components within that axis form the weight distribution with the sum of $1$. Then, the prototypical representation $Z_P \in \mathbb{R}^{n \times d}$ is obtained by element-wise multiplication between $Z_s \in \mathbb{R}^{n \times k \times d}$ and $W_{avg} \in \mathbb{R}^{n \times k \times 1}$, then summarize values along the $k$ axis to achieve the weighted mean representation on each category, which is the prototypical representation.

In our implementation, instead of feeding a fixed scalar into a linear layer, we directly initiate a learnable parameter with value $a = 1$ that shares the same dimensional space $\mathbb{R}^{n \times k \times d}$ and fed it to a softmax function for $W_{avg}$. We used the initial value of $1$ for the learnable parameter as we want to begin with averaging equally among feature vectors within the same category at first, but then weight values $w \in W_{avg}$ will vary through the training process to obtain a better prototypical representation that fits best with given data. In the experiment, the improved learnable prototypical embedding helps us achieve a better representation than the standard one and thus comes with higher performances in many scenarios.

\subsection{Cosine Transformer}
\subsubsection{Cosine similarity between two matrices}
Given two vectors $\mathbf{a}, \mathbf{b} \in \mathbb{R}^n$, the cosine similarity score between $\mathbf{a}$ and $\mathbf{b}$ is calculated by the dot-product of two vectors divided by the product of their magnitudes, by:
\begin{equation}
    S_C(\mathbf{a}, \mathbf{b}) = \frac{\mathbf{a} \cdot \mathbf{b}}{\|\mathbf{a}\| \cdot \|\mathbf{b}\|} = \frac{\sum_{i=1}^n a_i \cdot b_i}{\sqrt{\sum_{i=1}^n a_i^2}.\sqrt{\sum_{i=1}^n b_i^2}},
    \label{equa:Cosine_vectors}
\end{equation}

From the formula for vectors above, we expanded the definition of cosine similarity on matrices. Specifically, the cosine similarity $S_C(A, B) \in \mathbb{R}^{n \times m}$ between matrix $A\in \mathbb{R}^{n \times k}$ and $B \in \mathbb{R}^{k \times m}$ is \textit{the Hadamard division} between the matrix multiplication of $A$ and $B^\top$ and the outer product between vectors $\text{\textit{M}}_A \in \mathbb{R}^n$ and $\text{\textit{M}}_{B^\top} \in \mathbb{R}^m$, where $\text{\textit{M}}_A$ and $\text{\textit{M}}_{B^\top}$ are the vectorization of \textit{the magnitude values of the row vectors} of two matrices $A$ and $B^\top$, respectively. The definition of cosine similarity on two matrices is described by:

\begin{alignat}{2}
    S_C(A, B)&= (A \cdot B) &&\oslash (\text{\textit{M}}_A \otimes \text{\textit{M}}_{B^\top}) \label{equa:Cosine_matrices}\\
    &= (A\cdot B) &&\oslash
    \bigg(\begin{bmatrix}\sqrt{\sum_{i=1}^k a_{1,i}^2}, \cdots, \sqrt{\sum_{i=1}^k a_{n,i}^2}\ \ \end{bmatrix} \nonumber \\
    & && \otimes \begin{bmatrix}\sqrt{\sum_{i=1}^k b_{i,1}^2}, \cdots, \sqrt{\sum_{i=1}^k b_{i,m}^2} \ \ \end{bmatrix}\bigg),
    \nonumber
\end{alignat}

With that definition, individual element $S_C(A, B)_{i,j}$ is \textit{the cosine similarity score} between the row vector $a_i$ of matrix $A$ and the column vector $b_j$ of matrix $B$, where $i \in[1,\ n], \ j \in [1,\ m] $.

\subsubsection{Cosine attention mechanism}
Initially, as $n$ and $q$ are different in values, we reshaped the proto-feature $Z_P \in \mathbb{R}^{n \times d}$ into the $1 \times n \times d$ tensor and $Z_Q \in \mathbb{R}^{q \times d}$ into a $q \times 1 \times d$ tensor so we can maintain both $n$ and $q$ dimensions in the attention output $h_a \in \mathbb{R}^{q \times n \times d}$, providing the similarity matrix between $Z_P$ and $Z_Q$. With the two reshaped tensors, a set of three representations $\langle \mathbf{q^\ast},\ \mathbf{k},\ \mathbf{v} \rangle$ are obtained by linear layers:  $ \mathbf{q^\ast} = Z_P\cdot \theta_{q^{\ast}}^\top; \ \ \mathbf{k} = Z_{Q}\cdot \theta_k^\top; \ \ \mathbf{v} = Z_{Q}\cdot \theta_v^\top$ where $\theta_{\mathbf{q^\ast}}, \ \theta_\mathbf{k}, \ \theta_\mathbf{v} \in \mathbb{R}^{d \times d_h}$ are the weight matrices, $d_h$ is the dimension inside attention. The output attention head $h_a$ can be computed using the scaled dot-product or \textit{``Softmax Attention''} ($Soft\_Attn$) by:
\begin{align}
h_a^{Soft\_Attn} &= \mathcal{A} \cdot \mathbf{v} \nonumber \\ &= \textnormal{softmax}\left[(\mathbf{q^\ast} \cdot \mathbf{k}^\top) / \sqrt{d}\right] \cdot \mathbf{v},
    \label{equa:softmax_attn}
\end{align}

Specifically, in the softmax attention, the matrix multiplication performs dot-product operation between every pair of feature vectors between $\mathbf{q^\ast}$ and $\mathbf{k}^\top$, then divided by a scaling factor $\sqrt{d}$ before feeding to a softmax function for an attention map $\mathcal{A} \in \mathbb{R}^{q \times n \times 1}$, then multiplies with $\mathbf{v} \in \mathbb{R}^{q \times 1 \times d }$ for the attention output $h_{a} \in \mathbb{R}^{q \times n \times d}$. However, as there is an extra dimension for both $\mathcal{A}$ and $\mathbf{v}$, the matrix multiplication becomes the Hadamard product under tensor broadcasting. Therefore, in nature, the attention output $h_{a}$ becomes the element-wise multiplication between query features and the attention map of query samples and support proto-representations. While it is quite different from the usual concept, the nature of the attention mechanism is still preserved in $h_{a}$.

The core of the Softmax attention is the \textit{dot product operation}, which calculates the similarity involving vector angle and length (magnitude). The involvement of feature-length comes with limitations. First, the feature magnitude is a distinguishing yet unimportant factor in calculating attention, and the similarity output map can become unstable when feature magnitudes vary. This can be critical when $\mathbf{q}$ and $\mathbf{k}$ come from two distributions: few-shot support and query sets respectively. Second, the enlarging in magnitude makes the softmax function produces an extremely small gradient output, thus leading to gradient vanishing \cite{vaswani2017attention}. While the division of the fixed scaling factor $\sqrt{d}$ in softmax attention helps counter this phenomenon, the differential in feature magnitudes still remains. 

Therefore, to remove the effect of the differences in vectors' magnitude, we replaced the dot-product with cosine similarity for calculating $\mathcal{A}$. We clarify that the concept of replacing the dot-product operation with cosine similarity is not new and has been studied in \cite{luo2018cosine} and applied in few-shot image classification in \cite{chen2019closerfewshot, gidaris2018dynamic} in terms of a cosine similarity-based classification/recognition model. In this study, the replacement of cosine similarity in the attention mechanism helps us highlight the alignment between two representations by features' content. Specifically, instead of using a fixed number for scaling the entire weight matrix, individual components of the multiplicated matrix will be divided with the product of their corresponding vector's magnitude. We refer to this attention mechanism as \textit{``Cosine Attention''} $(Cos\_Attn)$ by Equation \ref{equa:cosine_attn}, based on the definition of cosine similarity for matrices as in Equation \ref{equa:Cosine_matrices}.

\begin{equation}
    h_a^{Cos\_Attn} = \left [(\mathbf{q^\ast} \cdot \mathbf{k}^\top) \oslash (\text{\textit{M}}_\mathbf{q^\ast} \otimes \text{\textit{M}}_\mathbf{k}) \right] \cdot \mathbf{v} ,
    \label{equa:cosine_attn}
\end{equation}

With cosine similarity, the attention map $\mathcal{A}$ focuses more on the features' content and determines a better correlation matrix between every pair of features $\langle q^\ast_i \in \mathbf{q^\ast}, \ k_j \in \mathbf{k}^\top \rangle$. Furthermore, the output distribution of cosine attention can be stable even if its input magnitude varies \cite{luo2018cosine}. By removing the magnitude, the output of cosine similarity is bounded into the range of $[-1,\ 1]$, indicating the similarity between the two features. Thus, the Softmax function is no longer necessary for scaling the values. Without the softmax, attention map $\mathcal{A}$ still maintains the probabilistic distributions in the row vectors as well as its components' ratio. Moreover, as cosine similarity does not scale the weight distributions into the sum of 1, $a_{i,j} \in \mathcal{A}$ possesses a wider range of value. This helps emphasize query features on aligned categories in $h_a$ and vice versa, hence boosting the model's performances. In our empirical experiment, removing softmax operation in cosine attention helps increase our proposed FS-CT performance significantly, and normalizing $\mathbf{q}$ and $\mathbf{k}$ before feeding to the Softmax attention does not procedure an attention map as adequate as using the cosine attention alone.

In our FS-CT method, we apply the multi-head mechanism for cosine attention. The initial three linear layers split $Z_P$ and $Z_Q$ into sets of $\langle \mathbf{q^\ast}_t,\ \mathbf{k}_t,\ \mathbf{v}_t \rangle$ where $t \in [1, 8]$. For each set, a corresponding attention output $h_a^t$ is computed by either softmax or cosine attention, represented for the projection output in different perspectives. Then, the $H_{out} \in \mathbb{R}^{q \times n \times d}$ is obtained with the output weight matrix $\theta_{\circ} \in \mathbb{R}^{d_h \times d}$ by:

\begin{equation}
    H_{out} = \textnormal{concat}(h_a^t)\cdot \theta_{\circ}^\top, \qquad t = 1,\cdots,8.
    \label{equal:multi_head}
\end{equation}

\subsubsection{Cosine linear layer for queries prediction}

After the attention block, two skip connections are performed on $H_{out} = (Z_{P} \ +\ H_{out}) \ +\ \textnormal{FFN}(Z_{P} \ +\ H_{out})$ with layer normalization before each step. The feed-forward network FFN is a simple two linear layers with GELU \cite{hendrycks2016gaussian} activation function in between. With the final outcome feature $H_{out} \in \mathbb{R}^{q \times n \times d}$, a linear layer with weight $\theta_{out} \in R^{d \times 1}$ is applied follows by softmax for $P_{out} \in \mathbb{R}^{q \times n}$, which represent the probabilistic prediction for every query features on $n$ categories. Instead of using a conventional linear layer, we used a cosine linear layer from \cite{chen2019closerfewshot}. Furthermore, instead of performing the dot-product between $H_{out}$ and $\theta_{out}$, cosine similarity $S_C(a,b)$ is replaced between two L2-normalized tensors with Equation \ref{equa:Cosine_vectors}. The replacement of cosine similarity instead of the convention dot-product operation helps us achieve a better prediction score for $P_{out}$. Overall, the probabilistic prediction $p(c\mid h_{q,c};\ \theta_{out})$ for representation score $ h_{q,c}$ on label $c$ of query sample $\boldsymbol{x}_q$  over $n$ categories and the predicted label $\hat y$ are calculated by:
\begin{align}
    p(c\mid h_{q,c};\ \theta_{out}) &= \frac{\text{exp}[S_C( h_{q,c},\ \theta_{out}^\top)]}{\sum_{i=1}^n \text{exp}[S_C( h_{q,i},\ \theta_{out}^\top)]},  \\
     \nonumber\\
    \hat y &=  \underset{c}{\arg\max}\ p(c\mid h_{q,c},\ \theta_{out}),
    \label{equal:prediction}
\end{align}

\begin{algorithm}[ht!]
{\fontsize{10pt}{10pt}\selectfont
\SetAlgoLined
\SetKwInOut{Input}{Input}
\SetKwInOut{Output}{Output}
\caption{\textbf{Episodic training algorithm of FS-CT} over one training epoch with $N$ tasks (episodic batch). Each task $\mathcal{T}_i$ is chosen randomly from the training set $D_\text{train}$ with a different set of categories to train and update the general parameter $\theta$ of our proposed FS-CT, including the embedding backbone $f(.\mid \theta_f)$.}
\label{algo:Episodic_Learning}
\vspace{.5em}
\Input{Training set $D_\text{train}$. The number of learning tasks $N$.\\ Learnable parameters $\theta$, learning rate $\theta$ }
\Output{Updated parameters $\theta$}
\vspace{.75em}
\For{$i \ \textnormal{in} \ \{1,...,N\}$}{
\nl Randomly chosen task $\mathcal{T}_i = \{S,\ Q\} \sim p(D_\textnormal{train})$. \;
\nl Obtain feature representations $Z_S,\ Z_Q$ from $S$ and $Q$ by the backbone $f(.\mid\theta_f)$\;
\nl Calculate the \textit{prototypical representation} $Z_P$ using Equation (\ref{equal:Weight_Averaging}). \;
\nl Obtain $\langle \mathbf{q^\ast}, \ \mathbf{k}, \ \mathbf{v} \rangle$ for attention by: \\$ \qquad  \mathbf{q^\ast} \gets Z_P\cdot \theta_{q^{\ast}}^\top; \quad \mathbf{k} \gets Z_{Q}\cdot \theta_k^\top; \quad \mathbf{v} \gets Z_{Q}\cdot \theta_v^\top$ \;
\nl Calculate $H_{out}$ by \textit{Cosine Attention} by Equations (\ref{equa:cosine_attn}) and (\ref{equal:multi_head}).\;
\nl Perform two skip-connections and FFN for \textit{Cosine Transformer}: \\$ \qquad  H_{out} \gets (Z_{P} \ +\ H_{out}) \ +\ \textnormal{linear}(\textnormal{GELU}[\textnormal{linear}(Z_{P} \ +\ H_{out})]) $ \;
\nl Obtain the query prediction scores $\hat{y}$ with Equation (\ref{equal:prediction}). \; 
\nl Compute the loss function $\pazocal{L}$ as in Equation (\ref{equa:loss_func}). \;
\nl Perform gradient descent step to update the parameter: \\$ \qquad \theta \gets \theta - \alpha \nabla_{\theta} \pazocal{L}$ \;
} 
}
\end{algorithm}

\subsection{Episodic Training}

We train our proposed FS-CT method with an episodic learning strategy, presented in detail in Algorithm \ref{algo:Episodic_Learning}. For each training step, task $\mathcal{T} = \{S,\ Q\}$ are selected randomly from $D_\text{train}$ with $n$ categories. For convenience, all learning parameters are referred to as a general parameter $\theta$. With FS-CT, includes backbone feature extraction $f(.\mid\theta_f)$, is performed, we obtain the probabilistic prediction $p(y\mid\boldsymbol{x}_Q, S; \theta)$ of sample $x_Q \in Q$ on label $y$ given $S$. Finally, Categorical Cross-entropy loss is applied to update parameters at the end of each training step as in Equation \ref{equa:loss_func}. 

\section{Experiment Results}
\label{sec:Experiment}
\subsection{Datasets and Experimental setup}
To evaluate our method, we adopt three standardized few-shot image classification datasets \textit{mini-ImageNet} \cite{vinyals2016matching}, CIFAR-FS \cite{bertinetto2018meta}, and CUB-200 \cite{wah2011caltech}. The \textit{mini-ImageNet} dataset is the subset of ImageNet dataset (ILSVRC-2012) \cite{russakovsky2015imagenet} that consists of 100 different categories with 600 image samples per each, each image having the size $84 \times 84$ pixels. In our implementation, we used the splits by Ravi and Laroche \cite{ravi2016optimization} including 64 training categories, 16 validation categories, and 20 testing categories. \textit{CIFAR-FS} including 100 categories containing 600 images for each label with the size of $32 \times 32$ pixels. The splits of this dataset are similar to \textit{mini-ImageNet}. The \textit{CUB-200} dataset contains 200 categories of bird species with 11,788 images of $84 \times 84$ pixels, which are divided into 100 categories for training, 50 categories for validating, and 50 categories for testing.

\begin{figure}[ht!]
    \centering
    \includegraphics[width=.8\linewidth]{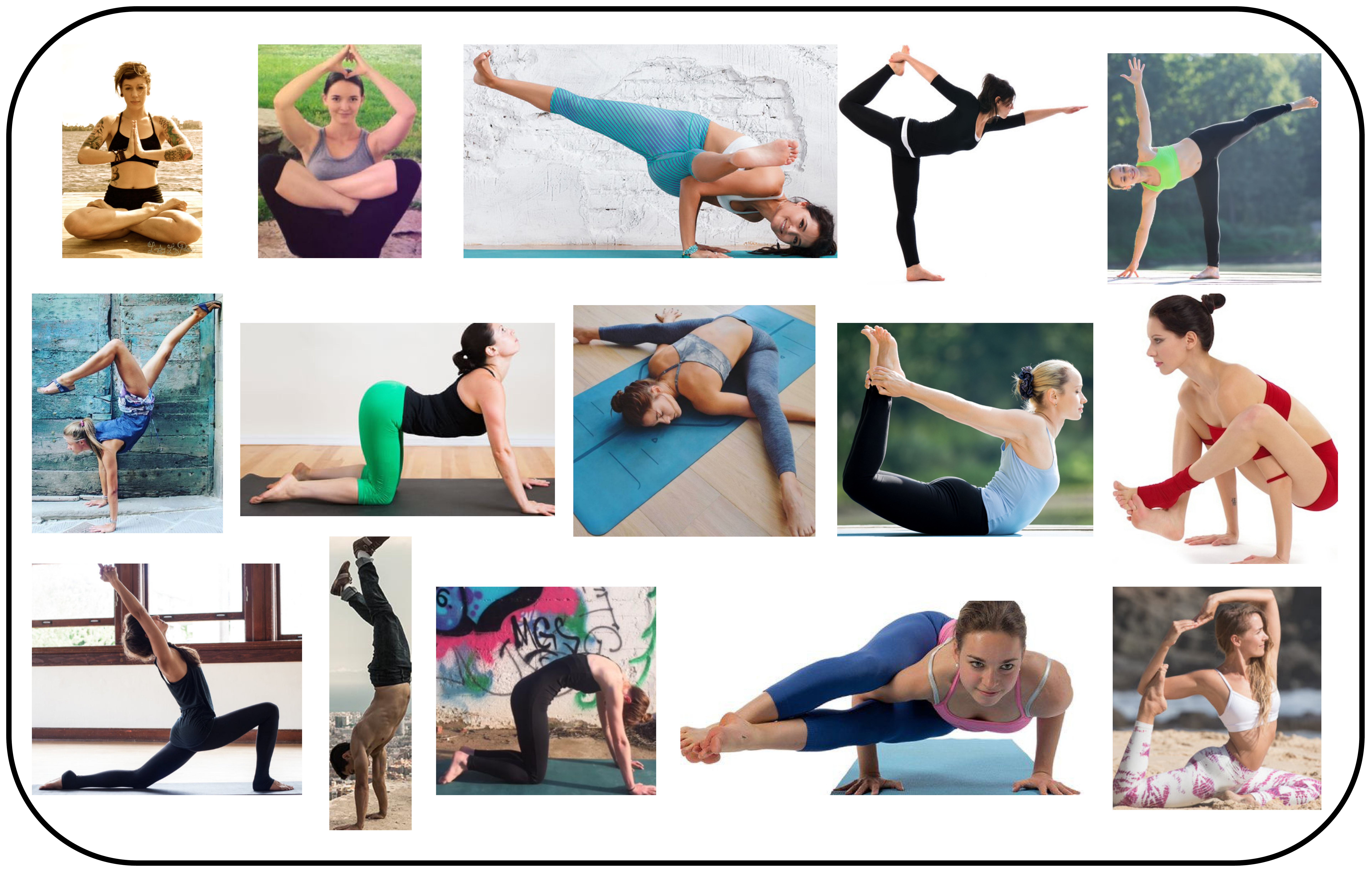}
    \caption{\textbf{Several exemplary samples for categorial poses of the yoga poses dataset}, consisting of 50 different poses with a total of 2,480 images.}
    \label{fig:Yoga_dataset}
\end{figure}

Besides the three few-shot datasets above, we also created a custom dataset for yoga poses scoring, including 50 categories of main yoga poses with 2,480 images. We developed this small-scale dataset as an initial step toward making a smart monitoring and study scheme for yoga participants. The dataset is partially derived from Kaggle \cite{saxena2021YogaKaggle} and stored in our implementation code on GitHub\footnote{\url{https://github.com/vinuni-vishc/Few-shot-transformer/tree/main/dataset/Yoga}}. The number of images are ranging from 30 to 81 samples per category with arbitrary size. Furthermore, some categories' samples are different in viewpoint, appearance, or visual condition, which makes the dataset more challenging. We split the dataset into 25 categories for training, 13 categories for validating, and 12 categories for testing. Some example poses from the dataset are presented in Fig. \ref{fig:Yoga_dataset}, with the statistics of the dataset distribution presented in Table \ref{table:Yoga_dataset}. 

\begin{table}[htbp]
\centering
\fontsize{8pt}{10pt}\selectfont
\renewcommand{\arraystretch}{1.5} 
    \caption{\textbf{Statistical description of the custom image dataset for Yoga poses} over three main sets for training, validation, and testing few-shot image classification method.}
    \begin{tabular}{lcccc}

    \hline
        & Train & Val & Test & \textbf{Total}\\
    \hline
    Number of Categories & 25 & 13 & 12 & \textbf{50}\\
    Min number of samples per category  & 30 & 39 & 31 &\textbf{30} \\
    Max number of samples per category  & 80 & 81 & 67 &\textbf{81} \\
    Average of samples per category & 50.2 & 49.9 & 48.0 & \textbf{49.6} \\
    Total samples & 1,255 & 649 & 576 & \textbf{2,480}\\
    \hline
    \end{tabular}
    \label{table:Yoga_dataset}
 \end{table}

 \begin{table*}[!t]
\centering
\fontsize{8pt}{10pt}\selectfont
\renewcommand{\arraystretch}{1.5} 
\caption{\textbf{Performance of our proposed \textit{FS-CT} for 5-way setting} on \textit{mini}-ImageNet, CUB-200, and CIFAR-FS, using either the baseline softmax attention (\textcolor{green}{Soft\_Attn}) or the proposed cosine attention (\textcolor{purple}{Cos\_Attn}) with two shallow backbones \textbf{Conv4} and \textbf{Conv6} with 50 training epoch and \textcolor{blue}{data augmentation}. Similar to \cite{chen2019closerfewshot}, we report the mean of 600 randomly generated test episodic tasks. The best and second best results are \textbf{bolded} and \underline{underlined}, respectively. The evaluation metric is accuracy in percentage.}
\begin{adjustbox}{max width=1.\textwidth,center}
    \begin{tabular}{lcc|cc||cc|cc||cc|cc}
        \hline
        & \multicolumn{4}{c||}{\textbf{mini-ImageNet}} & \multicolumn{4}{c||}{\textbf{CUB-200}} & \multicolumn{4}{c}{\textbf{CIFAR-FS}} \\
        \cline{2-13}
        \multicolumn{1}{c}{\multirow{2}{*}{\textbf{Methods} }} & \multicolumn{2}{c|}{\textbf{Conv4}} & \multicolumn{2}{c||}{\textbf{Conv6}} & \multicolumn{2}{c|}{\textbf{Conv4}} & \multicolumn{2}{c||}{\textbf{Conv6}} & \multicolumn{2}{c|}{\textbf{Conv4}} & \multicolumn{2}{c}{\textbf{Conv6}} \\
        & \textbf{1-shot} & \textbf{5-shot} & \textbf{1-shot} & \textbf{5-shot} & \textbf{1-shot} & \textbf{5-shot} & \textbf{1-shot} & \textbf{5-shot} & \textbf{1-shot} & \textbf{5-shot} & \textbf{1-shot} & \textbf{5-shot} \\
        \hline
        \textbf{FS-CT} + \textcolor{green}{Soft\_Attn} (\textit{baseline})                                & \underline{42.55} & 58.08 & 39.11 & 56.52 & 53.44 & 66.14 & \underline{55.41} & 65.87 & \underline{48.57} & 66.22 & \underline{46.66} & 65.43\\
        \textbf{FS-CT} + \textcolor{green}{Soft\_Attn} + \textcolor{blue}{Aug}         & 38.11 & 57.14 & 33.89 & 54.26 & 51.50 & \underline{69.73} & 48.61 & \underline{71.62} & 40.95 & 64.31 & 43.74 & 60.41\\
        \textbf{FS-CT} + \textcolor{purple}{Cos\_Attn}  (\textit{proposed})                              & \textbf{45.69} & \textbf{60.38} & \textbf{40.52} & \textbf{59.85} & \textbf{56.41} & 67.12 & 55.27 & 66.90 & \textbf{53.86} & \textbf{69.03} & \textbf{51.04} & \textbf{68.76}\\
        \textbf{FS-CT} + \textcolor{purple}{Cos\_Attn} + \textcolor{blue}{Aug}         & 40.73 & \underline{60.21} & \underline{35.33} & \underline{57.81} & \underline{56.34} & \textbf{75.88} & \textbf{55.68} & \textbf{76.93} & 47.85 & \underline{68.35} & 45.06 & \underline{67.40} \\
        \hline \\
    \end{tabular}
    \end{adjustbox}
    \label{table:main_result_conv}
\end{table*}

\begin{table*}[!ht]
\centering
\fontsize{8pt}{10pt}\selectfont
\renewcommand{\arraystretch}{1.5} 
\caption{\textbf{Performance of the baseline \textit{CTX} \cite{doersch2020crosstransformers} and our proposed \textit{FS-CT} for 5-way setting} on three datasets \textit{mini}-ImageNet, CUB-200, and CIFAR-FS, using the baseline softmax attention (\textcolor{green}{Soft\_Attn}) or the proposed cosine attention (\textcolor{purple}{Cos\_attn}), with two embedding backbones \textbf{ResNet-18} and \textbf{ResNet-34}, pre-trained on ImageNet.  \textcolor{blue}{Data augmentation} is applied for our FS-CT method only. Similar to Table \ref{table:main_result_conv}, we validate the methods with 600 random task episode tasks and report the mean value. The best and second best results are \textbf{bolded} and \underline{underlined}, respectively. The evaluation metric is accuracy in percentage.}{
    \begin{threeparttable}
    \begin{adjustbox}{max width=1.\textwidth,center}
    \begin{tabular}{lcc|cc||cc|cc||cc|cc}
        \hline
        & \multicolumn{4}{c||}{\textbf{mini-ImageNet}\tnote{*}} & \multicolumn{4}{c||}{\textbf{CUB-200}} & \multicolumn{4}{c}{\textbf{CIFAR-FS}} \\
        \cline{2-13}
        \multicolumn{1}{c}{\multirow{2}{*}{\textbf{Methods} }} & \multicolumn{2}{c|}{\textbf{ResNet-18}} & \multicolumn{2}{c||}{\textbf{ResNet-34}} & \multicolumn{2}{c|}{\textbf{ResNet-18}} & \multicolumn{2}{c||}{\textbf{ResNet-34}} & \multicolumn{2}{c|}{\textbf{ResNet-18}} & \multicolumn{2}{c}{\textbf{ResNet-34}} \\
        & \textbf{1-shot} & \textbf{5-shot} & \textbf{1-shot} & \textbf{5-shot} & \textbf{1-shot} & \textbf{5-shot} & \textbf{1-shot} & \textbf{5-shot} & \textbf{1-shot} & \textbf{5-shot} & \textbf{1-shot} & \textbf{5-shot} \\
        \hline
        \textbf{CTX} (\textit{baseline})  + \textcolor{green}{Soft\_Attn}   & 62.37 & 63.09 & 65.08 & 65.66 & 61.24 & 62.12 & 64.28 & 66.02 & 41.16 & 43.05 & 44.54 & 48.01\\
        \textbf{CTX} + \textcolor{purple}{Cos\_Attn}                                 & \underline{77.21} & \textbf{92.18} & \textbf{82.09} & \underline{93.41} & 73.14 & 85.71 & 76.54 & 88.15 & 60.18 & 73.97 & 61.82 & 74.83 \\
        \textbf{FS-CT} (\textit{our}) + \textcolor{green}{Soft\_Attn}                              & 72.36 & 84.13 & 73.90 & 85.58 & 74.69 & 85.38 & 75.97 & 87.03 & 63.40 & 77.98 & 63.05 & 78.04\\
        \textbf{FS-CT} + \textcolor{purple}{Cos\_Attn}  (\textit{proposed})                              & 75.93 & 90.12 & 79.01 & 91.64 & \underline{77.72} & \underline{89.13} & \underline{77.72} & \underline{89.63} & \underline{64.29} & \underline{80.57} & \underline{65.67} & \underline{81.44}\\
        \textbf{FS-CT} + \textcolor{green}{Soft\_Attn} + \textcolor{blue}{Aug}       & 73.84 & 84.65 & 74.61 & 88.74 & 75.89 & 88.06 & 77.01 & 89.05 & 62.41 & 78.17 & 62.14 & 79.17\\
        \textbf{FS-CT} + \textcolor{purple}{Cos\_Attn} + \textcolor{blue}{Aug}       & \textbf{77.40} & \underline{91.33} & \underline{80.32} & \textbf{93.82} & \textbf{81.12} & \textbf{91.96} & \textbf{81.23} & \textbf{92.35} & \textbf{64.49} & \textbf{81.46} & \textbf{67.06} & \textbf{82.89}\\
        \hline
    \end{tabular}
    \end{adjustbox}
    \begin{tablenotes}
    \item[*] For \textit{mini}-ImageNet, we will have a different experiment using \\more suitable pre-trained backbones for a fair comparison. 
    \end{tablenotes}
    \end{threeparttable}}
    \label{table:main_result_resnet}
\end{table*}

 \subsection{Implementation Details}
 We implemented our method and conducted experiments on PyTorch, using a CPU Intel Core i9-10900X 3.7GHz with a GPU NVIDIA GeForce RTX 3090 24GB and 16GB RAM memory. Methods were trained and experimented with the learning rate 0.001 without modification scheduler, AdamW \cite{loshchilov2017decoupled} optimization function with weight decay $1 \times 10^{-5}$, and no dropout. The model is optimized by Categorical Cross-entropy Loss Equation \ref{equa:loss_func}. These hyper-parameters are fixed as we want to make a fair comparison between all experiment scenarios. We performed two configurations: 5-way 5-shot and 5-way 1-shot, with 16 query samples for each category, making a total of 80 queries. All training steps are trained on 50 training epochs with 200 episodic batches (episodes) for each. Each training epoch is followed by a validating step with 200 episodes to select the best-performed model for the testing phase on 600 episodes. All training, validating, and testing sets are disjoined in categories. To increase training data samples for training models, we applied augmentation, including random resizing, cropping, horizontal flipping, color jittering, and image normalizing. All experiments were conducted in the same common ground of code, setting, and environment for a fair evaluation and comparison.

For backbone feature extraction, we mainly utilized four backbone models Conv4, Conv6, ResNet-18, and ResNet-32 \cite{he2016deep} for the experiments. Conv4 and Conv6 are lightweight CNN models with 4 and 6 layers respectively and trained from scratch without pre-training. These models have been used in previous studies on few-shot classification \cite{vinyals2016matching, chen2019closerfewshot, sung2018learning}. On the other hand, the ResNet backbone networks and pre-trained on mini-ImageNet are available on Torchvision. However, as the mini-ImageNet is a subset of the ImageNet, we deployed a special pre-trained model named FETI \cite{bateni2022enhancing} for evaluating the dataset, which will be described in detail later. For each type of backbone architecture, we resized sample images before training, depending on the dataset and backbone model. Particularly, with CNN backbones, we resize images into $64 \times 64$ pixels for CIFAR-FS (due to its small size originally) and $84 \time 84$ pixels for other datasets, and with Res-Net backbones, the resized input image is $112 \times 112$ and $224 \times 224$, respectively. For all experiments, we report results with Accuracy Equation \ref{equa:Acc} in percentage as the sole metric and use this performance metric for comparisons. The official implementation of our Few-shot Cosine Transformer and all experimental configurations are presented on our GitHub\footnote{\url{https://github.com/vinuni-vishc/Few-Shot-Cosine-Transformer}}.

\begin{equation}
    \text{Accuracy} = \frac{\text{Number of correct predictions}}{\text{Total number of predictions}} \times 100 ,
    \label{equa:Acc}
\end{equation}

\subsection{Ablation study}

\subsubsection{Evaluation on mini-ImageNet, CIFAR-FS, and CUB-200 datasets}
For the experiment, besides the FS-CT model, we also deploy another attention-based few-shot learning algorithm CTX \cite{doersch2020crosstransformers} as a baseline for the comparison with our proposed FS-CT method. We utilized both two attention mechanisms: the baseline Softmax attention Equation \ref{equa:softmax_attn} and our proposed cosine attention Equation \ref{equa:cosine_attn} for two few-shot methods for our ablation evaluation. The two main experiment results on mini-ImageNet, CIFAR-FS, and CUB-200 datasets are presented in Table \ref{table:main_result_conv} for the FS-CT method only with two embedding backbones Conv4 and Conv6, and Table \ref{table:main_result_resnet} for both CTX and FS-CT methods with ResNet-18, and ResNet-32 backbones. Both experiments are conducted in 5-way 1-shot and 5-way 5-shot settings. We use the full ImageNet pre-trained models on ResNet backbones for Table \ref{table:main_result_resnet} and we will have a separate experiment for a fair evaluation on mini-ImageNet in the latter section. Data augmentation is applied for our proposed FS-CT method only.

\begin{figure}[!hb]
    \centering
    \includegraphics[width=\linewidth]{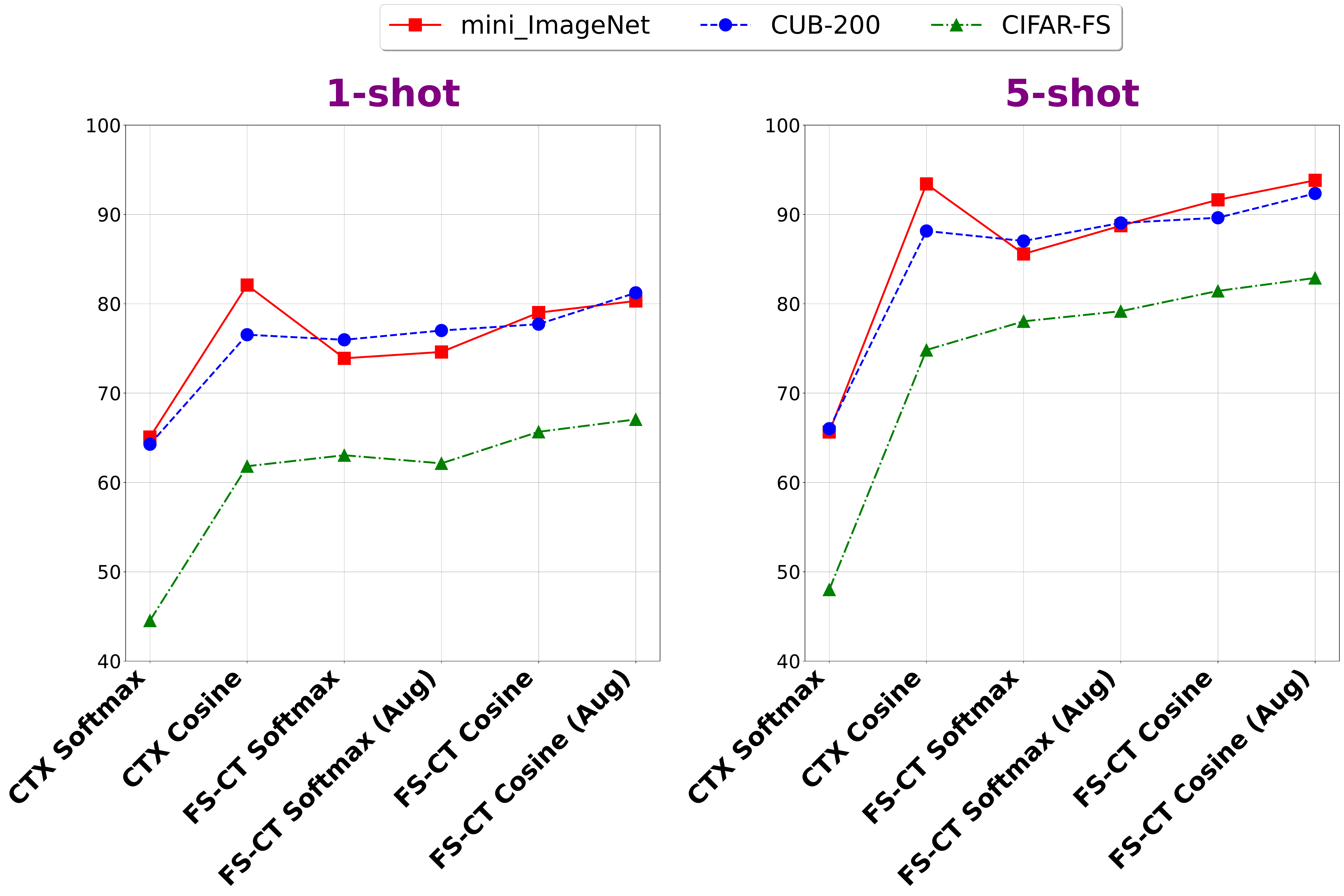}
    \caption{\textbf{Testing accuracies between CTX and FS-CT across the three few-shot datasets with softmax and cosine attention mechanisms and augmentation for FS-CT}, using ResNet-34 as the backbone. Generally, FS-CT achieved higher accuracies than CTX, with the only exception in mini-ImageNet, where the performances of CTX are nearly equal (in 5-shot learning) or higher (in 1-shot learning) than FS-CT. However, CTX only achieves these performances with our cosine attention mechanism as the core. Cosine attention improves the performances of two few-shot methods, and data augmentation further enhances them.}
    \label{fig:testing_plot}
\end{figure}

In general, our FS-CT outperformed CTX in different ResNet backbones, datasets, and few-shot settings in most cases with both two softmax attention and cosine attention, with the only exceptions on mini-ImageNet, where CTX outperformed FS-CT on 5-shot learning with ResNet-18 backbones and 1-shot learning with ResNet-34 backbones (Table \ref{table:main_result_resnet}). However, in both two cases, CTX is embedded with our proposed cosine attention mechanism, rather than the original version with softmax attention. On the other hand, cosine attention supports few-shot algorithms in outperforming the standard Softmax mechanisms across all backbones, few-shot settings, and datasets, with the improved performances increasing from nearly 5\% to over 20\% across cases. Overall, \textit{5-shot learning comes with better performance than 1-shot learning}. This happens typically on few-shot algorithms as more label samples come with better centroid representation for individual categories, thus classifying queries better. Furthermore, \textit{augmentation helps improve classifier performances} on FS-CT, mainly on the 5-shot setting. In particular, in Table \ref{table:main_result_resnet}, the second-best and best results mainly are FC-CT using cosine attention and its corresponding method training with augmentation, respectively. There are some occasions when augmentation does not help improve performance in one-shot learning. This could be explained by augmentation that comes with the growth in noise in categorical representation, therefore affecting the performances on one-shot learning. Augmentation seems to be effective on ResNet backbones, as FS-CT using cosine attention with augmentation mainly achieves the best result within individual scenarios. Across scenarios, \textit{deeper backbones comes with better performances}, as increasing the number of layers helps both CNN and ResNet backbones achieve higher results. Moreover, the models' performance is heavily affected by the choice of backbone, as in most cases, ResNet-34 backbone as feature extractor comes with the highest performance among the four. These observations are further illustrated in Fig. \ref{fig:testing_plot}, where the line graphs present the test accuracy correlations between CTX and FS-CT variants on ResNet-34 backbone across few-shot settings and datasets in Table \ref{table:main_result_resnet}.  

\begin{figure}[ht!]
\centering
    \includegraphics[width=.8\linewidth]{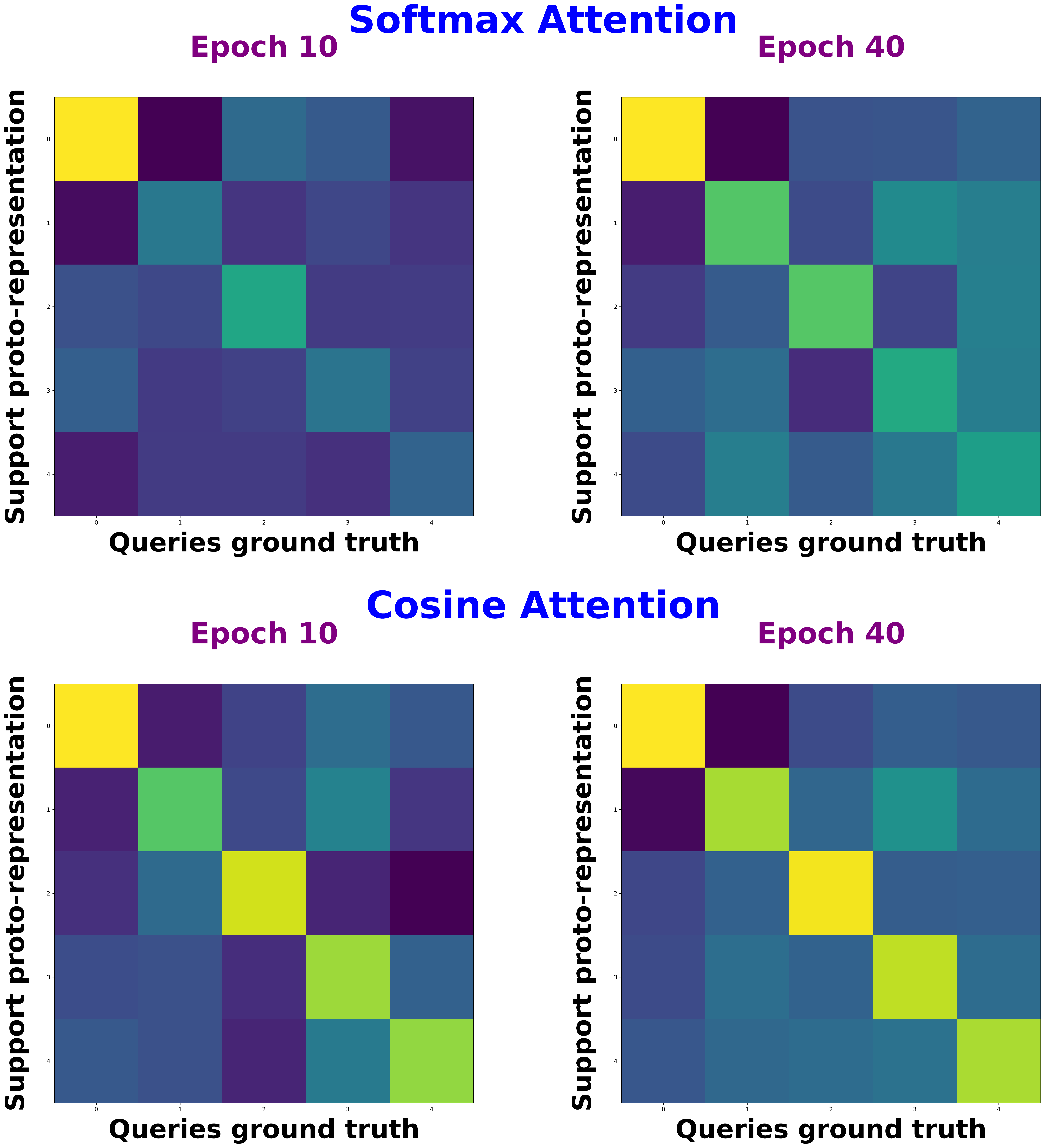}
    \caption{\textbf{Comparison between two attention outputs as the correlation heatmap of Softmax Attention (top row) and Cosine Attention (bottom row) on FS-CT, taken from epoch 10 (left column) and epoch 40 (right column) of the 50-epoch-training process}. The attention heatmap is between the query samples and support proto-representations that share the same ground truth categories. We use ResNet-34 as the backbone embedding and perform the 5-way 5-shot setting on the mini-ImageNet test set for the heatmap. Cosine attention produces more stable and robust attention heatmaps (through the main diagonal) from the early phase of training and becomes more apparent as the training progresses. This figure is better viewed in color.}
    \label{fig:attn_visual}
\end{figure}

\subsubsection{Performances of two attention mechanisms}

Fig. \ref{fig:attn_visual} illustrates the attention outputs as the correlation heatmap between the baseline Softmax Attention (top) and the improved Cosine Attention (bottom) for 5-way 5-shot learning, from the early training phase (left) to the later training phase (right). Each heatmap point represents the  similarity frequency between query samples (x-axis) from one category and the corresponding prototypical representation (y-axis) of the same category. All heatmaps are obtained on the performances from the same few-shot task derived from the testing test of mini-ImageNet. Cosine attention results in a more robust heatmap as it generates a stronger similarity matrix between the query and support samples (through proto-representations) that share the same ground truth, standing by the main diagonal. In the early epoch, the attention heatmap procedure by the cosine attention achieves a similar, if not better, than the Softmax attention heatmap from the latter epoch. The strong connection diagonal between queries and support representations with the same categorical ground truth becomes more apparent with cosine attention as training progresses, resulting in a more stable heatmap. This emphasizes the robustness of our improved cosine attention for the few-shot classification task.

\begin{figure*}[ht]
    \centering
    \includegraphics[width=\textwidth]{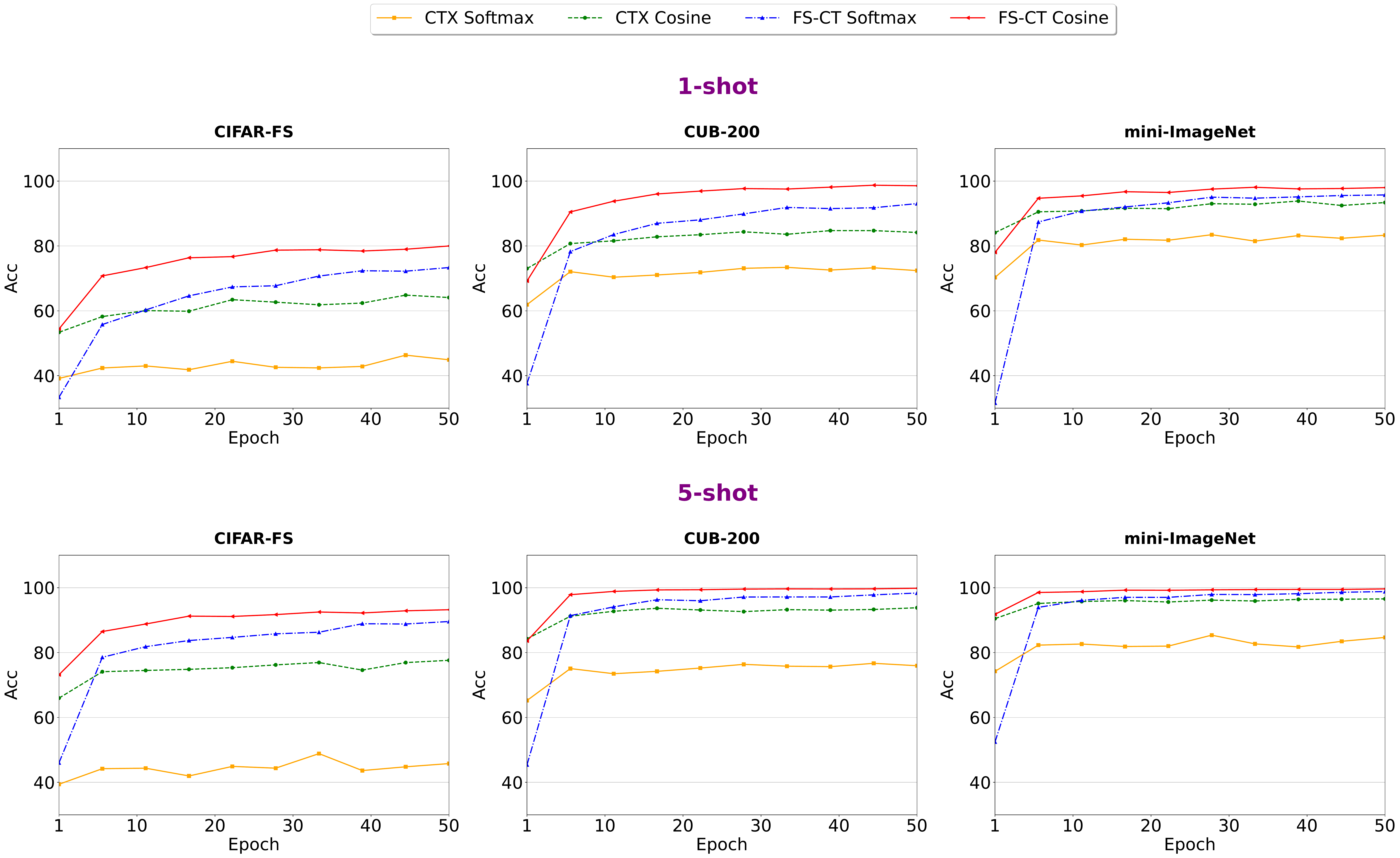}
    \caption{\textbf{Accuracy performances of FS-CT and CTX throughout training} for 1-shot and 5-shot setting with ResNet-34 backbone, using either the standard \textit{softmax attention} or the proposed \textit{cosine attention} mechanism. Cosine attention significantly improves both few-shot methods among settings and datasets with higher starting points and better training plots. FS-CT achieves better training performances than its counterpart CTX with both attention mechanisms.}
    \label{fig:training_plot}
\end{figure*}

Across Table \ref{table:main_result_conv} and  \ref{table:main_result_resnet}, we can observe that \textit{using cosine attention helps achieve more robust performance than Softmax attention}. This observation is further illustrated in Fig. \ref{fig:training_plot}, where cosine attention helps both methods acquire a higher accuracy at the starting point, and learn better training graphs across 50 epochs. The two methods' performances improve significantly as cosine attention produces a more stable attention outcome. Using cosine attention helps improve the learning ability consistently across datasets, settings, and methods compared with the baseline Softmax attention. Additionally, applying cosine attention without normalizing features is more effective than our early attempts to normalize two input feature sets before calculating the Softmax attention. This further highlights the robust improvement of our cosine attention for the cross-attention mechanism compared with the baseline scaled dot-product attention. Furthermore, the training graphs in Fig. \ref{fig:testing_plot} and Fig. \ref{fig:training_plot} also show that, under the same configuration (dataset, few-shot setting, backbone, and attention mechanism), our FS-CT learns and performs better than the baseline CTX significantly.

\subsubsection{Performances of FS-CT on mini-ImageNet with partially pre-trained model}

\begin{table}[htbp]
\centering
\fontsize{8pt}{10pt}\selectfont
\renewcommand{\arraystretch}{1.5} 
        \caption{\textbf{Performance of FS-CT on mini-ImageNet with a pre-trained model \textcolor{orange}{FETI}} (Feature Extractor Trained (partially) on ImageNet) that was trained with non-test-set overlapping ImageNet classes to avoid the natural advantage of ImageNet pre-trained model. We use two supported backbones ResNet-12 and ResNet-18 for the pre-trained model, with the same validating scheme and configurations.}
    \begin{tabular}{lcc|cc}
        \hline
        & \multicolumn{4}{c}{\textbf{mini-ImageNet}} \\
        \cline{2-5}
        \multicolumn{1}{c}{\multirow{2}{*}{\textbf{Methods} }} & \multicolumn{2}{c|}{\textbf{ResNet-12}} & \multicolumn{2}{c}{\textbf{ResNet-18}} \\
        & \textbf{1-shot} & \textbf{5-shot} & \textbf{1-shot} & \textbf{5-shot} \\
        \hline
        \textbf{FS-CT} (\textcolor{orange}{FETI}) + \textcolor{green}{Soft\_Attn}                                  & 45.70 & 57.03 & 40.06 & 58.20 \\
        \textbf{FS-CT} (\textcolor{orange}{FETI}) + \textcolor{purple}{Cos\_Attn}                                 & \textbf{55.62} & 70.36 & 41.84 & 71.34 \\
        \textbf{FS-CT} (\textcolor{orange}{FETI}) + \textcolor{green}{Soft\_Attn} + \textcolor{blue}{Aug}          & 38.88 & 57.16 & 47.72 & 62.71 \\
        \textbf{FS-CT} (\textcolor{orange}{FETI}) + \textcolor{purple}{Cos\_Attn} + \textcolor{blue}{Aug}         & 49.39 & \textbf{73.00} & \textbf{55.67} & \textbf{73.42} \\
        \hline
    \end{tabular}
    \label{table:mini_img_result}
\end{table}

As the mini-ImageNet dataset is a subset of the ImageNet dataset, using a default pre-trained ResNet on the full ImageNet as feature extraction comes with a naturally advanced performance on mini-ImageNet, as shown with very high performance in Table \ref{table:main_result_resnet}. Therefore, it is unfair for us if we want to compare our FS-CT performances with other few-shot classification methods. To tackle the problem, we used a specific pre-trained model that had been trained on a subset of ImageNet that is non-overlapped with the testing set of mini-ImageNet. We adapted this pre-trained model, called \textit{``Feature Extractor Trained (partially) on ImageNet''} or FETI in abbreviation, from \cite{bateni2022enhancing}. Because the pre-trained model was trained only on ResNet-18, we used two backbone models ResNet-18 and ResNet-12, as ResNet-32 is too different in layer number and layer size, and ResNet-12 is roughly adequate with ResNet-18 in architecture. Table \ref{table:mini_img_result} shows our FS-CT performance on mini-ImageNet with FETI pre-trained model. In general, the performances are reduced in the comparison with results from Table \ref{table:main_result_resnet}. This emphasizes the necessity of having a good embedding through a pre-trained model in order to address the few-shot problem, as pointed out in \cite{tian2020rethinking}. Results show that FS-CT with cosine attention still outperformed the baseline softmax attention, demonstrating the robust learning capability of our proposed attention mechanism.

\subsubsection{Evaluation on Yoga poses dataset}

\begin{table}[htbt]
\centering
\fontsize{8pt}{10pt}\selectfont
\caption{\textbf{Evaluation of the baseline CTX and our proposed FS-CT on the custom Yoga dataset} using two backbones ResNet-18 and ResNet-34 with 50 training epochs and 600 random testing tasks. The best and second best results are \textbf{bolded} and \underline{underlined}, respectively. The evaluation metric is accuracy in percentage.}
\renewcommand{\arraystretch}{1.5} 
    \begin{tabular}{lcc|cc}
        \hline
        & \multicolumn{4}{c}{\textbf{Yoga Dataset}} \\
        \cline{2-5}
        \multicolumn{1}{c}{\multirow{2}{*}{\textbf{Methods} }} & \multicolumn{2}{c|}{\textbf{ResNet-18}} & \multicolumn{2}{c}{\textbf{ResNet-34}} \\
        & \textbf{1-shot} & \textbf{5-shot} & \textbf{1-shot} & \textbf{5-shot} \\
        \hline
        \textbf{CTX} (\textit{baseline}) + \textcolor{green}{Soft\_Attn}   & 49.83 & 54.88 & 53.09 & 55.44 \\
        \textbf{CTX} + \textcolor{purple}{Cos\_Attn}                                 & 59.99 & \underline{76.61} & \underline{63.12} & \underline{77.80} \\
        \textbf{FS-CT} (\textit{our})  + \textcolor{green}{Soft\_Attn}                               & \underline{61.40} & 69.15 & 58.89 & 71.12 \\
        \textbf{FS-CT} + \textcolor{green}{Soft\_Attn} + \textcolor{blue}{Aug}       & 51.89 & 66.21 & 55.08 & 70.12 \\
        \textbf{FS-CT} + \textcolor{purple}{Cos\_Attn}                             & \textbf{66.38} & \textbf{80.34} & \textbf{64.32} & 77.66 \\
        \textbf{FS-CT} + \textcolor{purple}{Cos\_Attn} + \textcolor{blue}{Aug}       & 57.98 & 73.58 & 59.76 & \textbf{77.92} \\
        \hline
    \end{tabular}
    \label{table:Yoga_result}
\end{table}

For the custom Yoga poses dataset, the results are separately presented in Table \ref{table:Yoga_result}. While cosine attention still comes with more robust performances than the baseline Softmax attention, augmentation overall seems not to help the method in improving the outcome results. The best and second-best performances mainly are FS-CT and CTX, both with cosine attention, while training with augmentation resulted in much lower performances in FS-CT. Our theory is that this phenomenon is affected by the significant difference in hard cases and visual variation of the same pose category in the dataset. Still, these performance results showcase the potential of few-shot learning algorithms in general and our proposed FS-ST in particular for practical applications on healthcare topics, leading to our future studies in developing smart monitoring and scoring system for Yoga learners on downstream devices such as smartphones.

\section{Discussion}
\label{sec:Discussion}
We have proposed a transformer-based method for a few-shot classification task with an attention mechanism using cosine similarity. We find the algorithm of our proposed Few-shot Cosine Transformer (FS-CT, along with the improved cosine attention, is straightforward and simple to implement, with the detailed code implementation and experimental configurations presented in the previous section. Our experiments and ablation studies indicate that cosine similarity benefits the attention mechanism to produce a better and more consistent correlational map as attention output and enhance our framework performances across configurations, backbones, and few-shot settings. 

However, while our current results in various few-shot datasets are promising, there are some limitations that should be considered in future research. First, our method's performances highly depend on the choice of embedding backbone, particularly those with a pre-trained model. While the pre-trained backbones provide good embedding representation that supports few-shot learning algorithms to perform significantly, the dependence of the pre-trained model and its impact should be further investigated in future studies. Second, the complexity of architecture may prevent FS-CT from reaching higher performance levels. Although skip-connection was used to preserve information, it is possible that this was insufficient. Many few-shot approaches, including those from \cite{vinyals2016matching, chen2019closerfewshot, sung2018learning, snell2017prototypical} share a straightforward pipeline but perform well across few-shot datasets. We want to continue this line of research. Third, while we consider our improved learnable prototypical embedding is a simple method to address the support variation challenge and hard samples in few-shot learning, further exploration should be conducted in further studies to shed light on our improvement, especially when the training and testing sets come from two disjoined domains. Moreover, we believe that more efficient improvements to the prototype network are yet to be discovered, and the balancing between hard and easy samples has not been fully investigated. Last, due to our limited resources, we are only able to perform the comparison between our proposed cosine attention to the standard scaled dot-product attention, neglecting recent other exemplars of the attention mechanism. We suggest future studies based on our work should consider a more comprehensive and wide comparison across recent variations of the attention mechanism for vision transformer-based algorithms (not just limited to few-shot learning or image classification). We leave these limitations and discussions for future studies.

\section{Conclusion}
\label{sec:Conclusion}
In this study, we introduce Few-shot Cosine Transformer (FS-CT), a lightweight and straightforward transductive learning method for the few-shot image classification task based on the prototypical network and vision transformer. We made two improvements to our framework: (i) learnable prototypical embedding to balance between easy and hard samples of the provided labeled support instances and (ii) cosine attention based on cosine similarity to compute correlational map between support and query samples for few-shot recognition. Throughout extensive experiments and analysis, we prove that the cosine similarity supports the attention mechanism in providing a better and more consistent attention output as the correlational map, supporting FS-CT to achieve competitive results across few-shot datasets under various settings and configurations. The empirical results further show our proposed cosine attention also enhances the performances of other vision transformer-based few-shot algorithms as well. Finally, we showcase the potential of FS-CT in practical application in healthcare research via a custom yoga pose dataset. However, the potential of the proposed learnable prototypical embedding in dealing with hard samples and the impact of pre-trained models on few-shot learning algorithms should be investigated in further studies.

\bibliographystyle{splncs04}
\bibliography{reference}
\end{document}